\documentclass[11pt]{article}

\usepackage[final]{acl}

\usepackage{times}
\usepackage{latexsym}

\usepackage[T1]{fontenc}

\usepackage[utf8]{inputenc}

\usepackage{microtype}

\usepackage{inconsolata}

\usepackage{hyperref}
\usepackage{url}

\usepackage{hyperref}
\usepackage{url}
\usepackage{subcaption}
\usepackage{caption}
\usepackage{tabularx}
\usepackage{booktabs}
\usepackage{graphicx}
\usepackage{multirow}
\usepackage{wrapfig}
\usepackage{array}
\usepackage[utf8]{inputenc}
\usepackage{booktabs}
\usepackage{tabularx}
\usepackage{xcolor}
\usepackage{colortbl}
\usepackage{nicefrac}
\usepackage{amsmath}
\usepackage{amssymb}
\usepackage{makecell}
\usepackage{siunitx}
\usepackage[table]{xcolor}
\usepackage{algorithm}
\usepackage{algpseudocode}

\title{\textsc{Leash}: Adaptive Length Penalty and Reward Shaping \\ for Efficient Large Reasoning Model}

\author{
 \textbf{Yanhao Li\textsuperscript{1,$\ast$}},
 \textbf{Lu Ma\textsuperscript{1,$\ast$}},
 \textbf{Jiaran Zhang\textsuperscript{2}},
 \textbf{Lexiang Tang\textsuperscript{1}},
\\
 \textbf{Wentao Zhang\textsuperscript{1}},
 \textbf{Guibo Luo\textsuperscript{1,$\dagger$}},
\\
\\
 \textsuperscript{1}Peking University,
 \textsuperscript{2}Harbin Institute of Technology, Shenzhen, China,
\\
 \small{\textsuperscript{$\ast$}Equal contribution, \textsuperscript{$\dagger$}Corresponding author}
\\
 \small{
   \textbf{Correspondence:} \href{mailto:liyanhao@stu.pku.edu.cn}{liyanhao@stu.pku.edu.cn}, \href{mailto:luoguibo@pku.edu.cn}{luoguibo@pku.edu.cn}
 }
}

\begin{document}
\maketitle
\begin{abstract}
Large Language Models (LLMs) often produce unnecessarily lengthy reasoning traces, which significantly increase computational cost and latency. Existing approaches typically rely on fixed length penalties, but such penalties are hard to tune and fail to adapt to the evolving reasoning abilities of LLMs, leading to suboptimal trade-offs between accuracy and conciseness.
To address this challenge, we propose \textsc{Leash} \textit{(adaptive LEngth penAlty and reward SHaping)}, a reinforcement learning framework for efficient reasoning in LLMs. We formulate length control as a constrained optimization problem and employ a Lagrangian primal–dual method to dynamically adjust the penalty coefficient. When generations exceed the target length, the penalty is intensified; when they are shorter, it is relaxed. This adaptive mechanism guides models toward producing concise reasoning without sacrificing task performance.
Experiments on Deepseek-R1-Distill-Qwen-1.5B and Qwen3-4B-Thinking-2507 show that \textsc{Leash} reduces the average reasoning length by 60\% across diverse tasks—including in-distribution mathematical reasoning and out-of-distribution domains such as coding and instruction following—while maintaining competitive performance. Our work thus presents a practical and effective paradigm for developing controllable and efficient LLMs that balance reasoning capabilities with computational budgets.
\end{abstract}

\section{Introduction}
Large Reasoning Models (LRMs), driven by reinforcement learning (RL), have substantially improved reasoning capabilities by generating longer chains of thought (CoT) to solve complex reasoning tasks~\citep{deepseek_r1,openai-o1,openai-o3}. However, this enhancement often comes with \textit{token over-expansion} and \textit{redundant self-reflection}, leading to excessively long outputs and high computational costs~\citep{chen2025think23overthinkingo1like}.

To mitigate the issue of over-thinking, recent research has primarily focused on two directions: truncation-based CoT compression~\citep{hao2024traininglargelanguagemodels,ruan2025reasoninglearnlatentthoughts} and RL-based length-aware optimization~\citep{kimi1.5,aggarwal2025l1,hou2025thinkprune,laser,arora2025traininglanguagemodelsreason,su2025thinking}. The former directly limits the reasoning length by imposing a fixed generation budget, which effectively reduces computation but often cuts off essential intermediate reasoning steps, resulting in accuracy degradation. In contrast, RL-based methods explicitly incorporate length-aware objectives into the policy optimization process to dynamically balance accuracy and efficiency. For example, \textsc{ThinkPrune} enforces a strict generation-length cap during RL training, where tokens exceeding the limit are discarded~\citep{aggarwal2025l1}, while \textsc{L-CPO} formulates the task as a constrained optimization problem, introducing penalty terms to explicitly control the target generation length~\citep{hou2025thinkprune}. Although these methods demonstrate effectiveness in reducing overlong reasoning, they typically rely on fixed constraints or manually tuned penalty strengths, making them unable to adapt to the evolving reward distributions and reasoning dynamics during RL training. Consequently, they often suffer from unstable trade-offs: over-penalization suppresses necessary reasoning steps and harms accuracy, whereas under-penalization fails to constrain verbosity. This highlights the need for an adaptive mechanism that can dynamically adjust penalty strength according to real-time constraint satisfaction, enabling the model to automatically regulate reasoning length without external tuning.

To this end, we propose \textbf{\textsc{Leash}} (LEngth penAlty and reward SHaping), an RL-based adaptive length control method. \textsc{Leash} formulates reasoning-length control as a constrained optimization problem, aiming to maximize task reward while satisfying the expected length constraint. By introducing a dual variable $\lambda$ that adaptively adjusts the penalty strength based on the degree of constraint violation, \textsc{Leash} can flexibly ``tighten'' or ``loosen'' its ``leash'' on the output length—preserving critical reasoning steps while suppressing unnecessary verbosity.

To ensure training stability, we further design a one-sided penalized reward that penalizes only over-length generations while avoiding incentives for overly short outputs that could collapse model behavior. The dual variable $\lambda$ is updated via a primal–dual optimization mechanism, forming a feedback loop that automatically balances task reward and constraint satisfaction throughout training. Unlike fixed-constraint RL methods, this adaptive process continuously adjusts the degree of length control according to the evolving reasoning behavior, effectively preventing both over-penalization and under-penalization, and improving training stability and convergence.

We conduct comprehensive experiments on both mathematical and general reasoning benchmarks, including AIME24, AIME25, HMMT25, AMC23, GPQA, and MMLU-Pro, covering reasoning models of 1.5B and 4B scales. Experimental results show that \textsc{Leash} achieves a significantly better efficacy–efficiency trade-off than existing methods. On DeepSeek-R1-Distill-Qwen-1.5B, \textsc{Leash} reduces the average generation length by 62.7\% while improving accuracy by 0.8 points. Moreover, it maintains robust out-of-domain generalization on GPQA and MMLU-Pro. Training dynamics analysis reveals that \textsc{Leash} rapidly satisfies the length constraint in the early training stage and progressively stabilizes the dual coefficient $\lambda$, leading to smoother convergence and higher reasoning efficiency. Further behavioral analysis shows that the generated CoTs become more concise and focused, with redundant ``rethink'' or self-reflective phrases substantially reduced, while preserving essential planning and summarization structures.

\section{Method}
\begin{algorithm*}[!htbp]
\caption{\textsc{LEASH}: Adaptive Length Penalty and Reward Shaping}
\label{alg:leash}
\begin{algorithmic}[1]
\State \textbf{Input:} dataset $\mathcal{D}$, target length $L_t$, policy $\pi_\theta$
\State \textbf{Hyper:} $\alpha_\theta,\alpha_\lambda$, $[\lambda_{\min},\lambda_{\max}]$, $G$
\State Initialize $\theta$, $\lambda \ge 0$
\While{not converged}
  \State Sample prompts $\{x_b\}_{b=1}^B \sim \mathcal{D}$; set $\pi_\theta^{\text{old}} \gets \pi_\theta$
  \For{$b=1\to B$}
    \State Sample $G$ responses $y_{b,1:G}\sim \pi_\theta^{\text{old}}(\cdot|x_b)$ and lengths $L(y_{b,i})$
    \State Task reward $r_{b,i}\in\{-1,+1\}$ via $\operatorname{is\_equivalent}$
    \State One-sided penalty $\Delta_{b,i}\!=\!\max\!\bigl(0,\tfrac{L(y_{b,i})}{L_t}-1\bigr)$
    \State Shaped reward $\tilde r_{b,i}\!=\!\mathrm{clip}\bigl(r_{b,i}-\lambda\,\Delta_{b,i},-1,1\bigr)$
    \State Normalize advantages $\hat A_{b,i}\!=\!(\tilde r_{b,i}-\mu_b)/(\sigma_b\!+\!10^{-8})$
  \EndFor
  \State \textbf{Policy update:} maximize DAPO objective using $\hat A_{b,i}$; update $\theta \leftarrow \theta+\alpha_\theta\nabla_\theta J(\theta)$
  \State Estimate constraint violation $\,\widehat{J_P}\!=\!\frac{1}{BG}\sum_{b,i}\!(\tfrac{L(y_{b,i})}{L_t}-1)$
  \State \textbf{Dual update:} $\lambda \leftarrow \mathrm{clip}\!\bigl(\lambda+\alpha_\lambda\,\widehat{J_P},\,\lambda_{\min},\,\lambda_{\max}\bigr)$
\EndWhile
\State \textbf{return} $\pi_\theta$, $\lambda$
\end{algorithmic}
\end{algorithm*}
\textit{\textbf{Overview:}} We propose the adaptive \textbf{LE}ngth pen\textbf{A}lty and reward \textbf{SH}aping (\textbf{\textsc{Leash}}) method for efficient reasoning model training. This algorithm extends the DAPO~\cite{yu2025dapo} framework by introducing a Lagrangian constraint mechanism to effectively control the length of generated sequences while optimizing task performance. Concise pseudocode is provided in Algorithm~\ref{alg:leash}.

\noindent \textit{\textbf{Problem Formulation:}} To control generation length while maximizing the task objective, we formulate the problem as a constrained optimization task. The primary goal is to train a policy $\pi_\theta$ that maximizes the expected task-specific reward, while ensuring that the generation length $L(y)$ does not exceed a predefined target length $L_{\text{t}}$. The objective can be formally expressed as:
\begin{equation}
\begin{aligned}
\mathop{\text{maximize}}\limits_{\theta} \quad 
& \mathbb{E}_{x\sim\mathcal{D},\, y\sim\pi_\theta(\cdot|x)}[r(x,y)] \\
\text{s.t.} \quad 
& L(y) \le L_{\text{t}}, \quad 
\forall y \sim \pi_\theta(\cdot|x),
\end{aligned}
\label{equation:object}
\end{equation}
where $\mathcal{D}$ is a distribution of prompts, $y$ is the response generated by the LLM $\pi_\theta$. The task-specific reward function $r(x,y)$ provides a binary signal based on correctness, which utilizes a boolean function $\operatorname{is\_equivalent}(y, y^*)$, to determine if the generated response $y$ is semantically or functionally equivalent to the reference answer $y^*$:
\begin{equation}
\begin{aligned}
r(x,y) =
\begin{cases}
+1, & \text{if } \operatorname{is\_equivalent}(y,y^*) \text{ is true}, \\
-1, & \text{otherwise}.
\end{cases}
\end{aligned}
\end{equation}

However, the constraint in equation~\ref{equation:object} entails the challenge of guaranteeing length satisfaction for all possible responses, which is not straight forward to enforce directly with RL methods. In light of this, we reformulate the strict length constraint into an expectation form. This modification introduces a more tractable objective that aims to control the average generation length. Our surrogate objective is presented as follows:
\begin{equation}
\mathop{\text{maximize}}\limits_{\theta}\; J_{R}(\theta), \quad \text{s.t.} \quad J_P(\theta)\le 0,
\label{equation:JR&JP}
\end{equation}
where
\begin{equation}J_R(\theta) = \mathbb{E}_{x\sim\mathcal{D},\; y\sim\pi_\theta(\cdot|x)}\!\bigl[r(x,y)\bigr],
\end{equation}
\begin{equation}
J_P(\theta) = \mathbb{E}_{x\sim\mathcal{D},y\sim\pi_\theta(\cdot|x)}\left(\frac{L(y)}{L_\text{t}} - 1\right).
\end{equation}

To address this constrained problem, we leverage the Lagrangian method, a technique for finding the local maxima and minima of a function over a constraint set. This allows us to convert the constrained primal problem, as defined in equation~\ref{equation:JR&JP}, into its unconstrained saddle-point problem. We define the Lagrangian function $\mathcal{L}(\theta,\lambda)$ as:
\begin{equation}
\mathcal{L}(\theta, \lambda) = J_R(\theta) - \lambda \cdot J_P(\theta) ,
\end{equation}
where $\lambda \ge 0$ serves as the Lagrange multiplier. The goal is to find a saddle point $(\theta^*, \lambda^*)$ that solves the max-min problem:
\begin{equation}
    \mathop{\max}\limits_{\theta}\mathop{\min}\limits_{\lambda \ge 0} \mathcal{L}(\theta, \lambda).
\end{equation}

This equation encapsulates our primary goal, maximizing the task reward $J_R(\theta)$ while minimizing the length penalty $J_P(\theta)$, thereby encouraging the model to generate high-quality outputs of a controlled length. We use a Primal-Dual algorithm to find this saddle point by alternately updating the primal variable $\theta$ and the dual variable $\lambda$.

\noindent \textit{\textbf{Policy Update}}. In each iteration, we fix $\lambda$ and maximize $\mathcal{L}(\theta,\lambda)$ by performing one step of gradient ascent on $\theta$. Theoretically, the Lagrangian objective can be expressed as the expectation of an augmented reward signal:
\begin{equation}
\begin{alignedat}{2}
\mathcal{L}(\theta, \lambda)
&= J_R(\theta) - \lambda\, J_P(\theta) \\
&\hspace{-3.5em}
= \mathbb{E}_{x\sim\mathcal{D},\, y\sim\pi_\theta(\cdot|x)}
  \!\left[r(x,y) - \lambda\!\left(\frac{L(y)}{L_t} - 1\right)\right].
\end{alignedat}
\end{equation}
This formulation implies an augmented reward:
\begin{equation}
    r'(x,y) = r(x,y) - \lambda \cdot \left(\frac{L(y)}{L_t} - 1\right).
\end{equation}

However, directly optimizing this signal can be problematic: if a sequence is much shorter than the target $L(y) \ll L_t$, the term $- \lambda \cdot \left(\nicefrac{L(y)}{L_t} - 1\right)$ becomes a large positive bonus. This would incorrectly incentivize the policy to generate overly short responses, potentially harming task performance. We modify the reward signal for the policy update to be a one-sided penalized reward, which we denote as $r''(x,y)$. This practical adjustment ensures that we only penalize constraint violations without unintentionally rewarding conservative behavior, resulting in the modified reward $r''(x,y)$:
\begin{equation}
\begin{aligned}
r''(x,y)
= 
r(x,y)
- \lambda \cdot \max\!\left(0, \frac{L(y)}{L_t} - 1\right).
\end{aligned}
\end{equation}

Meanwhile, we also clip the reward signal to $[-1, 1]$ to prevent extreme rewards caused by fluctuations in the generated length $L(y)$ from triggering destructive gradient updates. The penalized reward is defined as:
\begin{equation}
\begin{aligned}
r'''(x,y)
= \text{clip}\!\left[r''(x,y),
-1,\, 1
\right].
\end{aligned}
\end{equation}

To maximize $r'''(x,y)$, we reframe this optimization problem as a standard RL task. We then apply the DAPO algorithm ~\cite{yu2025dapo}, using $r'''(x,y)$ as the reward signal to optimize policy loss $J(\theta)$. Notably, our approach deviates from the original DAPO implementation by omitting the dynamic sampling and overlong penalty techniques.


\begin{figure*}[!htbp]
    \centering
    \includegraphics[width=1\linewidth]{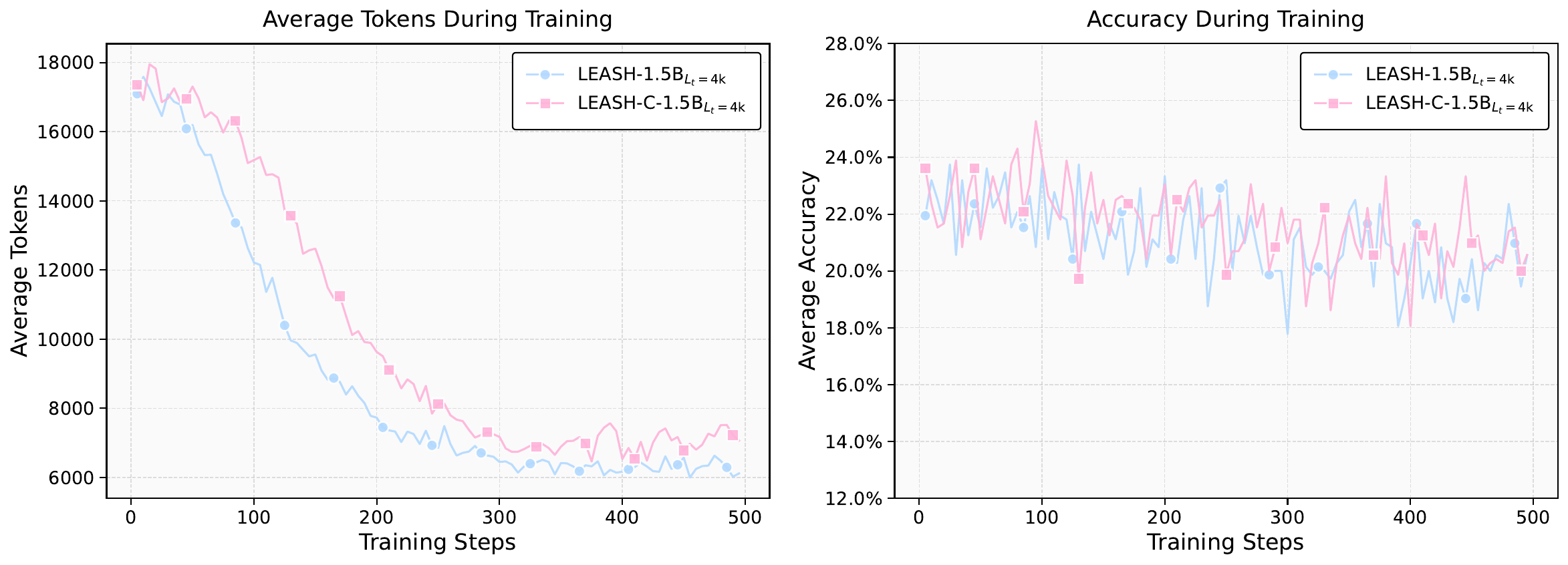}
    \caption{Training dynamics of \textsc{Leash} and \textsc{Leash}-C under the target length $L_t=4\mathrm{k}$ on the 1.5B base. Left: average tokens per response vs.\ training steps—\textsc{Leash} shortens trajectories faster and stabilizes at a lower length with smaller drift. Right: average accuracy vs.\ training steps—both variants remain comparable, indicating that length compression does not degrade task accuracy.}
    \label{fig:main-train}
\end{figure*}

\noindent \textit{\textbf{Dual Variable Update}}. After updating the policy parameters, we fix the policy parameters $\theta$ and update the dual variable. The update is performed via a gradient ascent step on the dual objective, which corresponds to adjusting $\lambda$ based on the constraint violation term $J_P(\theta)$.  The gradient of the Lagrangian with respect to $\lambda$ is:
\begin{equation}
\begin{aligned}
\nabla_\lambda \mathcal{L}(\theta, \lambda)
&= -J_P(\theta) \\
&\hspace{-2.5em} = -\mathbb{E}_{x\sim\mathcal{D},\, y\sim\pi_\theta(\cdot|x)}
   \left(\frac{L(y)}{L_\text{t}} - 1\right).
\end{aligned}
\end{equation}

Accordingly, the update rule applies this gradient. For practical stability, we also clip the resulting value to a predefined range $[\lambda_{min}, \lambda_{max}]$:
\begin{equation}
\begin{aligned}
\lambda_{k+1}
= \text{clip}\!\left[
\lambda_k + \alpha_\lambda \cdot J_P(\theta_{k+1}),
\, \lambda_{\min},\, \lambda_{\max}
\right].
\end{aligned}
\end{equation}

This update rule creates an adaptive adjustment mechanism for the penalty coefficient $\lambda$. The direction of the adjustment is determined by the sign of the constraint violation term $J_P(\theta_{k+1})$. If the average length exceeds the target $(J_P(\theta_{k+1})>0)$, $\lambda$ increases to intensify the penalty on length in the next policy update step. Conversely, if the average length is below the target $(J_P(\theta_{k+1})<0)$, $\lambda$ decreases to relax the penalty, shifting the optimization focus back to maximizing the primary objective $J_{R}$.

By alternating between these primal and dual updates, the algorithm dynamically adjusts both the policy and the penalty strength to find an equilibrium that maximizes the task reward while satisfying the average length constraint.

\section{Experiments}
\begin{table}[!htbp]
\centering
\resizebox{\linewidth}{!}{ 
\begin{tabular}{lcc}
\toprule
Variant Label & $\lambda$ Specification \\
\midrule
L\textsc{eash-C-1.5B}$_{L_t=\text{4k}}$ & $\lambda=0.1$ \\
L\textsc{eash-1.5B}$_{L_t=\text{4k}}$ & $\lambda_{\text{init}}=0.1, \lambda_{\text{lr}}=0.005$ \\
L\textsc{eash-C-4B}$_{L_t=\text{12k}}$ & $\lambda=0.2$ \\
L\textsc{eash-4B}$_{L_t=\text{12k}}$ & $\lambda_{\text{init}}=0.2, \lambda_{\text{lr}}=0.005$ \\
L\textsc{eash-C-4B}$_{L_t=\text{8k}}$ & $\lambda=0.2$ \\
L\textsc{eash--4B}$_{L_t=\text{8k}}$ & $\lambda_{\text{init}}=0.2, \lambda_{\text{lr}}=0.05$ \\
\bottomrule
\end{tabular}
}
\caption{Penalty settings for \textsc{Leash} variants at target lengths $L_t$. The suffix ``\textsc{-C}'' indicates a \emph{constant} penalty (fixed $\lambda$), while variants without ``\textsc{-C}'' use \emph{adaptive} dual updates with initialization $\lambda_{\text{init}}$ and step size $\lambda_{\text{lr}}$. Rows labeled \emph{1.5B} are based on \textit{DeepSeek-R1-Distill-Qwen-1.5B}, and rows labeled \emph{4B} are based on \textit{Qwen3-4B-Thinking-2507}. The table reports the exact $\lambda$ or $(\lambda_{\text{init}}, \lambda_{\text{lr}})$ used for each configuration.}
\vspace{-0.2cm}
\label{tab:lambda-schedules}
\end{table}
\begin{table*}[!htbp]
\definecolor{highlightcolor}{RGB}{255, 255, 204} 
\definecolor{goodchange}{RGB}{34, 139, 34}      
\definecolor{badchange}{RGB}{220, 20, 60}       
\centering
\scriptsize 
\setlength{\tabcolsep}{4pt} 
\renewcommand{\arraystretch}{1.1} 
\resizebox{\textwidth}{!}{
\begin{tabular}{l|rr|rr|rr|rr|cccc}
\toprule

\multirow{3}{*}{\textbf{Model}} & \multicolumn{2}{c|}{\textbf{AIME24}} & \multicolumn{2}{c|}{\textbf{AIME25}} & \multicolumn{2}{c|}{\textbf{HMMT25}} & \multicolumn{2}{c|}{\textbf{AMC}} & \multicolumn{4}{c}{\textbf{Overall Performance}} \\
\cmidrule(lr){2-3} \cmidrule(lr){4-5} \cmidrule(lr){6-7} \cmidrule(lr){8-9} \cmidrule(lr){10-13}

& \makecell{Avg.\\@32} & \makecell{Avg.\\Tokens} & \makecell{Avg.\\@32} & \makecell{Avg.\\Tokens} & \makecell{Avg.\\@32} & \makecell{Avg.\\Tokens} & \makecell{Avg.\\@32} & \makecell{Avg.\\Tokens} & \makecell{Avg.\\Acc.} & \makecell{Avg.\\Tokens} & \makecell{$\Delta$.\\Acc.} & \makecell{$\Delta$.\\Tokens(\%)} \\
\midrule
\multicolumn{13}{c}{\textit{Based on DeepSeek-R1-Distill-Qwen-1.5B}} \\
\midrule
Original Model & 31.4 & 16,722 & 23.1 & 16,562 & 14.5 & 18,521 & 63.3 & 11,103 & 33.1 & 15,727 & \multicolumn{2}{c}{\textit{Baseline}} \\
L1-Qwen-1.5B-Max & 28.3 & 3,140 & 19.6 & 2,948 & 11.7 & 3,074 & 68.2 & 2,413 & 32.0 & 2,893 & \textcolor{badchange}{-1.1} & \textcolor{goodchange}{-81.6\%} \\
L1-Qwen-1.5B-Extra & 27.9 & 2,843 & 20.4 & 2,577 & 8.3 & 2,597 & 69.1 & 2,226 & 31.4 & 2,561 & \textcolor{badchange}{-1.6} & \textcolor{goodchange}{-83.7\%} \\
\tiny{$\text{LASER-DE}_{L_T=4096}$} & 32.8 & 8,115 & 22.8 & 7,690 & 14.9 & 8,808 & 67.7 & 4,849 & 34.6 & 7,366 & \textcolor{goodchange}{+1.5} & \textcolor{goodchange}{-53.2\%} \\
$\text{T\textsc{hink}T\textsc{rune}}_{4k{\scriptsize\to}3k{\scriptsize\to}2k}$ & 25.9 & 5,999 & 21.1 & 5,419 & 12.5 & 6,607 & 65.9 & 3,451 & 31.4 & 5,369 & \textcolor{badchange}{-1.7} & \textcolor{goodchange}{-65.9\%} \\
L\textsc{eash-C-1.5B}$_{L_t=\text{4k}}$ & 27.9 & 7,587 & 23.0 & 6,883 & 11.3 & 7,094 & 63.7 & 4,976 & 31.5 & 6,635 & \textcolor{badchange}{-1.6} & \textcolor{goodchange}{-57.8\%} \\
L\textsc{eash-1.5B}$_{L_t=\text{4k}}$ & 30.4 & 6,779 & 24.6 & 6,126 & 14.2 & 6,358 & 66.4 & 4,230 & 33.9 & 5,873 & \textcolor{goodchange}{\textbf{+0.8}} & \textcolor{goodchange}{\textbf{-62.7\%}} \\
\midrule
\multicolumn{13}{c}{\textit{Based on Qwen3-4B-Thinking-2507}} \\
\midrule
Original Model & 80.8 & 19,193 & 74.6 & 21,414 & 52.9 & 24,645 & 93.8 & 12,961 & 75.5 & 19,553 & \multicolumn{2}{c}{\textit{Baseline}} \\
L\textsc{eash-C-4B}$_{L_t=\text{12k}}$ & 78.8 & 15,941 & 73.3 & 18,042 & 49.0 & 19,720 & 92.7 & 10,110 & 73.5 & 15,953 & \textcolor{badchange}{-2.1} & \textcolor{goodchange}{-18.4\%} \\
L\textsc{eash-4B}$_{L_t=\text{12k}}$ & 79.7 & 14,211 & 73.3 & 16,700 & 51.9 & 17,926 & 93.3 & 8,873 & 74.6 & 14,428 & \textcolor{badchange}{\textbf{-1.0}} & \textcolor{goodchange}{\textbf{-26.2\%}} \\
L\textsc{eash-C-4B}$_{L_t=\text{8k}}$ & 74.6 & 12,127 & 67.5 & 14,044 & 46.3 & 14,569 & 91.0 & 7,548 & 69.9 & 12,072 & \textcolor{badchange}{-5.7} & \textcolor{goodchange}{-38.3\%} \\
L\textsc{eash--4B}$_{L_t=\text{8k}}$ & 76.6 & 12,086 & 66.5 & 14,135 & 46.0 & 14,671 & 91.1 & 7,277 & 70.1 & 12,042 & \textcolor{badchange}{-5.5} & \textcolor{goodchange}{-38.4\%} \\
\bottomrule
\end{tabular}
}
\caption{
In-domain mathematical reasoning performance across four benchmarks (AIME24, AIME25, HMMT25, and AMC). 
Positive changes in accuracy and reductions in token usage are highlighted in \textcolor{goodchange}{green}, while degradations are marked in \textcolor{badchange}{red}. 
\textsc{Leash} consistently achieves the best efficacy--efficiency trade-off, substantially shortening reasoning trajectories while maintaining or improving accuracy compared with fixed-penalty (\textsc{Leash-C}) and prior baselines such as L1-Qwen-1.5B-Max/Extra, \textsc{ThinkPrune} and \textsc{LASER-DE}.
}
\label{tab:main_benchmark_iid}
\end{table*}
\subsection{Experimental Setup}
\paragraph{Models and Datasets.}
Representative open-source long–reasoning LLMs include 
DeepSeek-R1~\citep{deepseek_r1} and Qwen3-235B-A22B-Thinking~\citep{qwen3}, along with their distilled variants.  In our experiments, we adopt two representative models from these families: DeepSeek-R1-Distill-Qwen-1.5B\footnote{\url{https://huggingface.co/deepseek-ai/DeepSeek-R1-Distill-Qwen-1.5B}} and Qwen3-4B-Thinking-2507\footnote{\url{https://huggingface.co/Qwen/Qwen3-4B-Thinking-2507}}. For training, we construct a refined subset of the DAPO-MATH-17k dataset\footnote{\url{https://huggingface.co/datasets/BytedTsinghua-SIA/DAPO-Math-17k}}~\citep{yu2025dapo} by filtering out prompts where Qwen3-4B-Thinking-2507 consistently succeeds or consistently fails under pass@4 evaluation, yielding 3{,}939 instances used in our experiments.

\paragraph{Training configuration.}
The training is conducted using the VeRL~\citep{verl} framework’s DAPO training loop, with overlong reward shaping and dynamic sampling disabled. We train on 32 NVIDIA H20 GPUs with a global batch size of 64, using Adam optimizer with a learning rate of $1\times 10^{-6}$ and a sampling temperature of 1.0. KL regularization is not applied, and the entropy coefficient is set to 0.0. Rewards are clipped to the range $[-1, 1]$, with DAPO clipping thresholds set asymmetrically at $\epsilon_{\text{low}}=0.2$ and $\epsilon_{\text{high}}=0.28$, while the dual variable is bounded by $\lambda_{\min}=0.0$ and $\lambda_{\max}=1.0$. The maximum context length during training is 32k tokens, with 1k tokens reserved for the prompt.
\vspace{-0.2cm}

\paragraph{Baselines and Variants}
We include the original base checkpoints as baselines for comparison. Additionally, we re-evaluate the open-sourced checkpoints of L1-Qwen-1.5B-Max/Extra~\citep{aggarwal2025l1}, T\textsc{hink}P\textsc{rune}~\citep{hou2025thinkprune} and L\textsc{asert}-DE~\citep{laser} within our evaluation protocol. To investigate the effect of the length constraint and dual updates, we also train variants under different target lengths $L_t$. For clarity, we refer to the runs with a constant penalty as L\textsc{eash-C}, and those with the adaptive dual update as the L\textsc{eash} variant. The specific schedules for the penalty ($\lambda$) applied to each target $L_t$ are shown in Table~\ref{tab:lambda-schedules}. All other hyperparameters align with the training configuration outlined earlier.

\paragraph{Evaluation protocol.}
We evaluate our checkpoints on several widely used reasoning benchmarks, including AIME 24, AIME 25, HMMT 25~\citep{hmmt25} and AMC 23, with 32 samples per prompt for the first four datasets. We also include out-of-domain sets GPQA~\citep{gpqa} and MMLU-Pro~\citep{mmlu_pro} in our evaluation. The maximum generation length is set to 32,768 tokens, which includes both the reasoning budget tokens and final answer tokens, aligning with the training phase. Unless specified otherwise, we use a sampling temperature of $T=0.6$, top-$p=0.95$, and top-$k=-1$.

\section{Results and Analysis}
\subsection{Main Results}
We summarize the main experimental results of our method in Table~\ref{tab:main_benchmark_iid} and Table~\ref{tab:main_benchmark_ood}, which present the model performance on in-domain mathematical reasoning benchmarks and out-of-domain general reasoning tasks, respectively. Beyond the aggregate results, we also examine how \textsc{Leash} evolves during training. As shown in Figure~\ref{fig:main-train}, the adaptive variant reduces average tokens more quickly and settles at a lower plateau than the constant-penalty version, indicating faster and steadier length control.
\begin{table}[!htbp]
\centering
\definecolor{goodchange}{RGB}{34, 139, 34}      
\definecolor{badchange}{RGB}{220, 20, 60}       
\scriptsize 
\setlength{\tabcolsep}{4pt} 
\renewcommand{\arraystretch}{1.1} 
\resizebox{\linewidth}{!}{
\begin{tabular}{l|rr|rr|cccc}
\toprule

\multirow{3}{*}{\textbf{Model}} & \multicolumn{2}{c|}{\textbf{GPQA}} & \multicolumn{2}{c|}{\textbf{MMLU-Pro}} & \multicolumn{4}{c}{\textbf{Overall Performance}} \\
\cmidrule(lr){2-3} \cmidrule(lr){4-5} \cmidrule(lr){6-9}

& \makecell{Acc} & \makecell{Avg.\\Tokens} & \makecell{Acc} & \makecell{Avg.\\Tokens} & \makecell{Avg.\\Acc.} & \makecell{Avg.\\Tokens} & \makecell{$\Delta$.\\Acc} & \makecell{$\Delta$.\\Tokens(\%)} \\
\midrule

\multicolumn{9}{c}{\textit{Based on DeepSeek-R1-Distill-Qwen-1.5B}} \\
\midrule
Original Model & 17.2 & 10,943 & 18.9 & 6,099 & 18.0 & 8,521 & \multicolumn{2}{c}{\textit{Baseline}} \\
L\textsc{eash-C-1.5B}$_{L_t=\text{4k}}$ & 22.3 & 4,379 & 19.4 & 2,619 & 20.8 & 3,499 & \textcolor{goodchange}{+2.8} & \textcolor{goodchange}{-58.9\%} \\
L\textsc{eash-1.5B}$_{L_t=\text{4k}}$ & 22.7 & 3,908 & 19.7 & 2,222 & 21.2 & 3,064 & \textcolor{goodchange}{+3.2} & \textcolor{goodchange}{-64.0\%} \\

\midrule

\multicolumn{9}{c}{\textit{Based on Qwen3-4B-Thinking-2507}} \\
\midrule
Original Model & 65.8 & 9,082 & 74.0 & 4,742 & 69.9 & 6,912 & \multicolumn{2}{c}{\textit{Baseline}} \\
L\textsc{eash-C-4B}$_{L_t=\text{12k}}$ & 68.2 & 7,435 & 74.1 & 4,049 & 71.2 & 5,742 & \textcolor{goodchange}{+1.3} & \textcolor{goodchange}{-16.9\%} \\
L\textsc{eash-4B}$_{L_t=\text{12k}}$ & 66.7 & 7,182 & 74.1 & 3,900 & 70.4 & 5,541 & \textcolor{goodchange}{+0.5} & \textcolor{goodchange}{-19.8\%} \\
L\textsc{eash-C-4B}$_{L_t=\text{8k}}$ & 61.1 & 6,311 & 73.8 & 3,510 & 67.5 & 4,910 & \textcolor{badchange}{-2.4} & \textcolor{goodchange}{-29.0\%} \\
L\textsc{eash--4B}$_{L_t=\text{8k}}$ & 64.6 & 6,302 & 73.9 & 3,406 & 69.3 & 4,854 & \textcolor{badchange}{-0.6} & \textcolor{goodchange}{-29.8\%} \\
\bottomrule
\end{tabular}
}
\caption{Out-of-domain (OOD) general reasoning performance on GPQA and MMLU-Pro. Positive accuracy gains and reductions in token usage are highlighted in \textcolor{goodchange}{green}, while degradations are shown in \textcolor{badchange}{red}. We report per-benchmark accuracy and average tokens, along with overall averages and deltas. }
\vspace{-0.3cm}
\label{tab:main_benchmark_ood}
\end{table}
\paragraph{In-domain reasoning performance.}
As shown in Table~\ref{tab:main_benchmark_iid}, the proposed \textsc{Leash} series achieves a favorable efficacy--efficiency trade-off, substantially shortening the generation length while maintaining or even improving reasoning accuracy. 
For the 1.5B model, \textsc{Leash-1.5B}$_{L_t=\text{4k}}$ reduces the generation length by \textbf{62.7\%} while improving accuracy by \textbf{+0.8} points over the original model, clearly outperforming \textsc{ThinkPrune} and \textsc{Laser-DE} under the same length constraint.
For the 4B model, the adaptive dual-update variants demonstrate the best overall balance. 
Specifically, \textsc{Leash-4B}$_{L_t=\text{12k}}$ achieves a \textbf{26.2\%} reduction in output length while maintaining \textbf{74.6\%} average accuracy, only 1.0 points below the base model, showing stable performance even under aggressive constraints.
Across all experiments, adaptive variants consistently outperform the constant-penalty ones, confirming that dynamically adjusting the dual variable $\lambda$ effectively stabilizes training and prevents over-penalization.

\paragraph{Out-of-domain generalization.}
We further evaluate the models on GPQA and MMLU-Pro to assess generalization beyond the training distribution, as shown in Table~\ref{tab:main_benchmark_ood}. 
The results show that \textsc{Leash} significantly improves efficiency while maintaining competitive accuracy. 
On Qwen3-4B-Thinking-2507, \textsc{Leash-4B}$_{L_t=\text{12k}}$ improves the average accuracy by \textbf{+1.3 points} while reducing generation length by \textbf{16.9\%}, suggesting that moderate length constraints can enhance focus and consistency in reasoning. 
When the constraint is further tightened to 8k, \textsc{Leash-4B}$_{L_t=\text{8k}}$ still achieves about 29\% token reduction with only minor performance degradation, demonstrating strong robustness across domains.


\begin{figure*}[!htbp]
    \centering
    \includegraphics[width=1\linewidth]{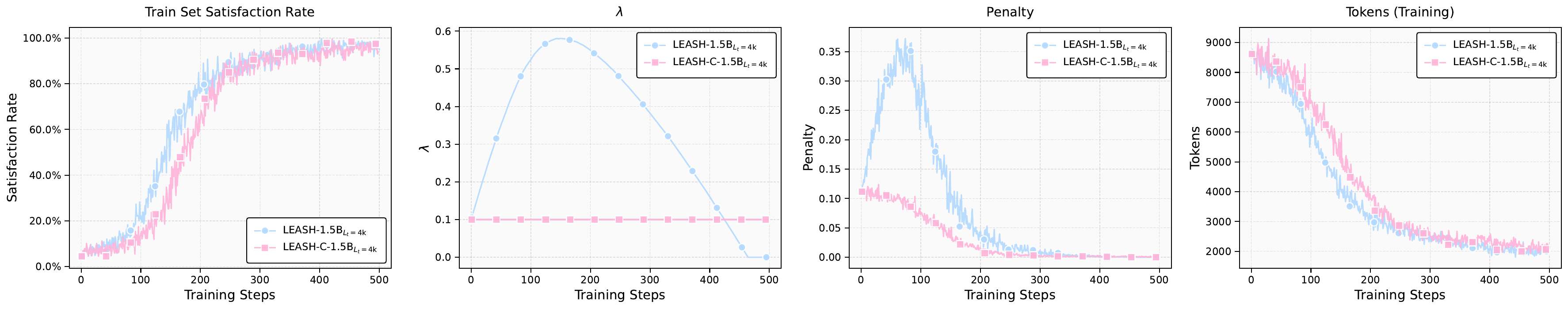}
    \caption{Training dynamics of \textsc{Leash} and \textsc{Leash}-C under $L_t = 4\text{k}$. The plots show (from left to right): (1) satisfaction rate on the training set, (2) adaptive penalty coefficient $\lambda$, (3) effective penalty value, and (4) average token length. \textsc{Leash} dynamically adjusts $\lambda$ to accelerate convergence and stabilize constraint satisfaction.}
    \label{fig:train_dynamics_leash}
\end{figure*}
\subsection{Behavior over Training}
We further analyze the training behaviors of \textsc{Leash} to better understand how the dual updates interact. Figure~\ref{fig:train_dynamics_leash} presents the training dynamics of the proposed \textsc{Leash} and its constant-penalty variant (\textsc{Leash}-C) under the target length $L_t = 4\text{k}$. Four key metrics are tracked: the satisfaction rate of the length constraint, the adaptive penalty coefficient $\lambda$, the effective penalty value, and the average number of generated tokens during training.

In the early phase of training (approximately the first 100 steps), \textsc{Leash} exhibits a sharp increase in satisfaction rate, indicating that the model quickly learns to generate responses within the target length range. During this period, since the average length initially exceeds the target, the penalty coefficient $\lambda$ gradually increases to enforce stronger constraint pressure. Around step 150, $\lambda$ reaches its peak and then steadily decreases as the model stabilizes near the target length. By approximately step 450, $\lambda$ approaches zero, suggesting that the model has effectively satisfied the constraint and no longer requires further penalty adjustment. The penalty term follows a similar pattern: it rises early in training as the system strengthens length control, then progressively decays as the generated outputs converge toward the desired token budget. The average token count decreases sharply from around 9k tokens to below 3k and eventually stabilizes near 2k by the end of training, corresponding to a reduction of more than 75\% relative to the initial generation length.

In contrast, the constant-penalty variant \textsc{Leash}-C maintains a fixed $\lambda$, resulting in a smoother but less adaptive penalty schedule. While both methods ultimately achieve near-perfect satisfaction rates (close to 100\%), the adaptive version converges faster and demonstrates more stable control. These observations highlight \textsc{Leash}’s ability to dynamically adjust its penalty coefficient in response to constraint violations, thereby achieving a stable balance between generation efficiency (shorter outputs) and constraint satisfaction.

\begin{figure*}[!htbp]
    \centering
    \includegraphics[width=1\linewidth]{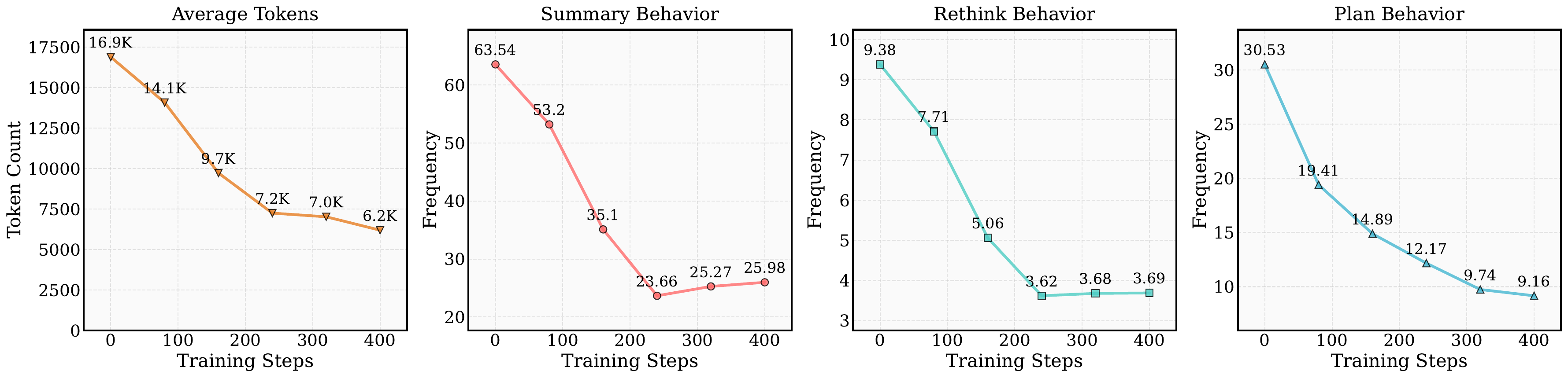}
    \caption{Evolution of the model’s thinking patterns across RL iterations on AIME2024 and AIME2025, using DeepSeek-R1-Distill-Qwen-1.5B as the base model. The average response length (left) consistently decreases throughout training, indicating progressive compression of reasoning trajectories. Correspondingly, the frequencies of summary-, rethink-, and plan-related keywords (right) exhibit distinct dynamics: summary and rethink behaviors decline sharply before stabilizing with slight rebounds, while plan behaviors continuously diminish.}
    \label{fig:behaviors}
\end{figure*}
\subsection{Changes of Thinking Patterns}
To better understand the mechanisms behind the changes in response length, we analyze the evolution of the model’s thinking patterns across RL iterations on the AIME2024 and AIME2025 datasets, as shown in Figure~\ref{fig:behaviors}. For each question, 16 samples are generated, revealing substantial shifts in the model’s reasoning dynamics during training. The average response length steadily decreases from 16.9K to 6.2K tokens, indicating that the model gradually learns to produce more concise reasoning trajectories while maintaining completeness. As the responses become shorter, different types of reasoning behaviors also exhibit systematic changes. \textit{Summary-related} keywords (e.g., ``So'', ``Therefore'', ``Thus'') show an overall decreasing trend, dropping from 63.5 to about 24, but slightly rising and stabilizing around 26 in the later stages. This suggests that the model rapidly reduces redundant summarization in the early phase while later reintroducing moderate summarization behaviors to maintain the completeness of conclusions. Notably, despite the slight increase in keyword frequency, the overall responses remain shorter, indicating that the model retains essential structural conclusions while achieving a balance between brevity and completeness. \textit{Rethink-related} expressions (e.g., ``recheck'', ``re-evaluate'') decrease from 9.38 to 3.62 and then slightly rebound to 3.69, showing that the model reduces excessive self-verification while preserving necessary reflective mechanisms. This aligns with our optimization objective, as the goal of the \textsc{Leash} algorithm’s RL training is to reduce redundant and inefficient reflection rather than suppress reflective reasoning entirely. The minor rebound indicates that the model reintroduces limited reflective behaviors after compression to maintain self-correction capability, achieving a balance between conciseness and robustness. Meanwhile, \textit{plan-related} markers (e.g., ``First'', ``Second'', ``Step'') consistently decrease from 30.5 to 9.2, suggesting that explicit step-by-step planning is gradually replaced by more implicit and compact procedural reasoning. We provide the full list of tracked keywords for each reasoning behavior category in Appendix~\ref{appendix:keywords}.

\section{Related Works}
\subsection{RL-Driven Emergence of Reasoning Abilities}
Reinforcement learning has recently emerged as a promising approach for enhancing the reasoning capabilities of LLMs. Early breakthroughs, such as OpenAI o-series~\cite{openai-o1,openai-o3}, DeepSeek R1~\cite{deepseek_r1}, and Kimi k-series~\cite{kimi1.5,kimik2}, have demonstrated that applying Reinforcement Learning with Verifiable Rewards (RLVR) can lead to significant improvements in complex reasoning tasks. 

Building on these foundations, Light-r1~\cite{Logic-rl} incorporates curriculum learning to refine reasoning skills, while Logic-rl~\cite{Logic-rl} adopted logic-driven reward functions to improve general reasoning ability. Deepscaler~\cite{Deepscaler} trains models with progressively longer contexts as their performance improves. Based on GRPO, DAPO~\cite{yu2025dapo} uses clip-higher to prevent entropy collapse, dynamic sampling for improved efficiency, and token-level policy gradient loss with overlong reward shaping to stabilize training. ~\cite{yue2025does, yan2025learning, ma2025learning} focus on introducing and addressing the concern that RLVR does not impart fundamentally new reasoning abilities to LLMs.

\subsection{Efficient Large Reasoning Models}
Recent large reasoning models (LRMs) have achieved remarkable performance on complex reasoning tasks by generating long CoTs, enabling effective problem-solving in domains such as mathematics and coding. However, while LRMs significantly improve performance on reasoning tasks, they also cause substantial overhead. Compared to LLMs, reasoning models lead to redundancy across multiple dimensions. Consequently, several methods have been proposed to mitigate this CoT length redundancy ~\cite{kimi1.5,feng2025efficient}.

Several studies investigate techniques to control response length during inference time. For example, ~\cite{hassid2025don} indicates that shorter reasoning chains tend to be more accurate, and suggests voting over the shortest m of k samples. Similarly, ~\cite{yang2025dynamic} observes that monitoring model behavior and dynamically terminating reasoning chains enhances both accuracy and efficiency. Additionally, ~\cite{muennighoff2025s1} introduces "budget forcing" employing specific phrases such as "Wait" or "Final Answer" to manage when reasoning ceases without the need for retraining. These methodologies complement LEASH and can be synergistically combined to reduce inference time costs.

Numerous studies focus on training models with shorter correct responses. Several works employ the concept of rejection sampling to fine-tune models, prompting shorter responses ~\cite{shrivastava2025sample,kim2024m,kimi1.5}. Concurrently, other studies introduce explicit length-aware penalties within the reward of RLVR to discourage excessively long reasoning chains. For example, ~\cite{yu2025dapo} employs reward shaping to encourage shorter responses; ~\cite{hou2025thinkprune} imposes a token limit during during RL, awarding zero rewards beyond this cap and progressively tightening it; ~\cite{su2025thinking} implements an adaptive direct length penalty that evolves over training to prevent over/or under-compression; ~\cite{xiang2025just} inversely scales the penalty with the prompt's solve rate; and ~\cite{aggarwal2025l1} optimizes accuracy while adhering to a prompt-specified target length by penalizing deviations. Although these methods empirically adopt length penalties, the degree of the penalty does not dynamically adjust according to the model's capabilities. Our method, LEASH, proposes a more flexible penalty strategy from a mathematical perspective, offering advantages in theory and in experiments.

\section{Conclusion}
We introduce \textsc{Leash}, a reinforcement learning framework that frames length control as a constrained optimization problem and solves it with a Lagrangian primal–dual scheme. A one-sided length penalty, clipped reward shaping, and an adaptive update of the dual variable $\lambda$ guide the policy toward concise and accurate reasoning without manual retuning. On the \emph{DeepSeek-R1-Distill-Qwen-1.5B} and \emph{Qwen3-4B-Thinking-2507} bases, \textsc{Leash} delivers a favorable efficacy–efficiency balance on in-domain mathematical reasoning, as summarized in Table~\ref{tab:main_benchmark_iid}, and it generalizes well to GPQA and MMLU-Pro, as reported in Table~\ref{tab:main_benchmark_ood}. The adaptive variant cuts tokens by roughly sixty percent on average while preserving, and at times improving, accuracy. Training dynamics in Figure~\ref{fig:train_dynamics_leash} show faster convergence than a constant-penalty counterpart, and the behavior study in Figure~\ref{fig:behaviors} indicates that \textsc{Leash} suppresses redundant summarization and self-reflection yet preserves essential structure. Overall, \textsc{Leash} offers a practical recipe for building controllable and compute-efficient reasoning models that respect token budgets.

\section{Limitations and Future Work}
\textsc{Leash} is simple and broadly applicable. Our study uses two compact backbones and standard benchmarks to keep compute modest and results reproducible. The limitations are mainly about scope. We plan to scale to larger models and to multi-turn/tool-use settings to test generality, and to replace a fixed $L_t$ with a learned or scheduled target that adapts to task difficulty and latency budgets, enabling automatic token allocation without sacrificing accuracy. We will also explore richer rewards and lightweight auto-tuning of $\lambda$ for stability.



\bibliography{acl_latex_main}

\appendix

\clearpage
\begin{table*}[!ht]
\centering
\small
\begin{tabular}{l|p{11cm}}
\toprule
\textbf{Behavior Category} & \textbf{Representative Keywords} \\
\midrule
\textbf{Summary-related} & 
\texttt{So}, \texttt{Therefore}, \texttt{Thus}, \texttt{conclude}, \texttt{overall} \\[4pt]
\textbf{Rethink-related} & 
\texttt{check again}, \texttt{double-check}, \texttt{re-evaluate}, \texttt{re-examine}, \texttt{reanalyze}, \texttt{reassess}, \texttt{recheck}, \texttt{reconsider}, \texttt{reevaluate}, \texttt{reevaluation}, \texttt{reexamine}, \texttt{rethink}, \texttt{think again}, \texttt{verify again}, \texttt{wait} \\[4pt]
\textbf{Plan-related} & 
\texttt{first}, \texttt{First}, \texttt{Second}, \texttt{second}, \texttt{step}, \texttt{Step} \\
\bottomrule
\end{tabular}
\caption{Keyword groups used for behavior analysis.}
\label{tab:keyword_groups}
\end{table*}

\section{Appendix}
\subsection{Keyword Groups for Thinking Behavior Analysis}
\label{appendix:keywords}
To analyze the evolution of thinking behaviors, we categorize representative keywords into three groups: \textit{Summary}, \textit{Rethink}, and \textit{Plan}, as shown in Table~\ref{tab:keyword_groups}. These keywords are matched case-insensitively within generated responses to compute their frequency across RL iterations.

\subsection{Use of LLM Assistants}
We used large language models (e.g., ChatGPT, Gemini) solely for language editing (grammar/clarity), light \LaTeX{} tweaks, and occasional debugging hints. 
All technical ideas, algorithms, experiments, and analyses were designed, implemented, and verified by the authors. 
No proprietary or personal data were shared with LLM services; only non-sensitive prompts about public benchmarks and settings were used. 
All results are reproducible from the released code and hyperparameters; LLM-assisted edits do not affect numerical outcomes.




\end{document}